\documentclass[conference]{IEEEtran}
\IEEEoverridecommandlockouts
\usepackage{cite}
\usepackage{amsmath,amssymb,amsfonts}
\usepackage{algorithmic}
\usepackage{graphicx}
\usepackage{textcomp}
\usepackage{xcolor}
\usepackage{url}
\usepackage{multirow}
\usepackage{graphicx}
\def\BibTeX{{\rm B\kern-.05em{\sc i\kern-.025em b}\kern-.08em
    T\kern-.1667em\lower.7ex\hbox{E}\kern-.125emX}}

\begin{document}

\title{WT-BCP: Wavelet Transform based Bidirectional Copy-Paste for Semi-Supervised Medical Image Segmentation}


\author{
Mingya Zhang$^{1,}$  \quad Liang Wang$^{1, *}$ \quad Limei Gu$^{2}$ \quad Tingsheng Ling$^{2}$ \quad Xianping Tao$^{1}$\\
    $^{1}$ Nanjing University \quad
    $^{2}$ Jiangsu Provincial Hospital of Traditional Chinese Medicine
}

\maketitle

\begin{abstract}
Semi-supervised medical image segmentation (SSMIS) shows promise in reducing reliance on scarce labeled medical data. 
However, SSMIS field confronts challenges such as distribution mismatches between labeled and unlabeled data, artificial perturbations causing training biases, and inadequate use of raw image information, especially low-frequency (LF) and high-frequency (HF) components.
To address these challenges, we propose a Wavelet Transform based Bidirectional Copy-Paste SSMIS framework, named WT-BCP, which improves upon the Mean Teacher approach. 
Our method enhances unlabeled data understanding by copying random crops between labeled and unlabeled images and employs WT to extract LF and HF details.
We propose a multi-input and multi-output model named XNet-Plus, to receive the fused information after WT. 
Moreover, consistency training among multiple outputs helps to mitigate learning biases introduced by artificial perturbations. During consistency training, the mixed images resulting from WT are fed into both models, with the student model's output being supervised by pseudo-labels and ground-truth. Extensive experiments conducted on 2D and 3D datasets confirm the effectiveness of our model.
Code: \url{https://github.com/simzhangbest/WT-BCP}.
\end{abstract}


\section{Introduction}
\label{sec:intro}

Medical image segmentation is a crucial task in medical image analysis and it plays an essential role in accurately classifying pixels in medical images to locate lesions~\cite{Wang2019abdominal}. Nevertheless, the segmentation performance is strongly dependent on the amount of training samples and requires labels from experienced physicians with clinical expertise~\cite{Li2018hdenseunet,Dou2020unpaired,Zhao20213d}. In order to reduce the reliance on labeled training data, many semi-supervised learning methods have great potential in making full use of the abundantly available massive unlabeled data in practice.

\begin{figure}[!htp]
    \centering
    \includegraphics[width=\linewidth]{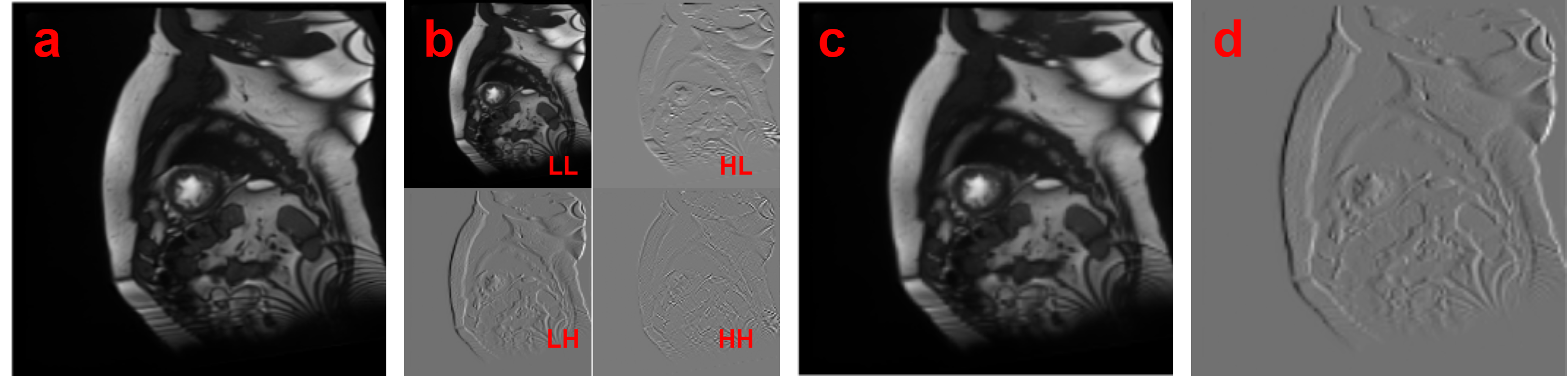}
    \caption{Take ACDC as an example, visualize LF and HF results. a. Raw image. b. Wavelet transform results. c. LF image. d. HF image.}
    \label{fig:wt_abc}
\end{figure}
CutMix~\cite{CutMix} is regarded as an effective and powerful data augmentation method, which has the potential to encourage unlabeled data to learn common semantics from the labeled data, since pixels in the same map share semantics to be closer \cite{Wang2022separated}. It uses patches of random sizes from the source image to replace areas of the same size in the target image. We have noticed that introducing random copy-pasting between labeled and unlabeled samples can create effective data disturbances in semi-supervised learning, leading to remarkable enhancements compared to multiple established baselines. 
BCP~\cite{bai2023bidirectional} redesigned the CutMix size and method of random copy-pasting. In the task of semi-supervised medical image segmentation, labeled and unlabeled data are bidirectionally copy-pasted. Through a large number of experiments, it is proved that random cropping 2/3 volum can achieve better results on 3D and 2D medical datasets.

Perform BCP processing on the input and directly hand it over to the Mean Teacher (MT)~\cite{ref1} framework to train and supervise. (1) Since the framework often adds input perturbations and applies consistency regularization between teacher and student, a single model structure inevitably produces noisy or erroneous pseudo-labels~\cite{ref2}, resulting in model confirmation and cognitive biases~\cite{arazo2020pseudo,ref4}. 
(2) Perturbations in consistency-based semi-supervised models are often artificially designed. They may introduce negative learning bias that are not beneficial for training.

\begin{figure*}[ht]
    \centering
    \includegraphics[width=\linewidth]{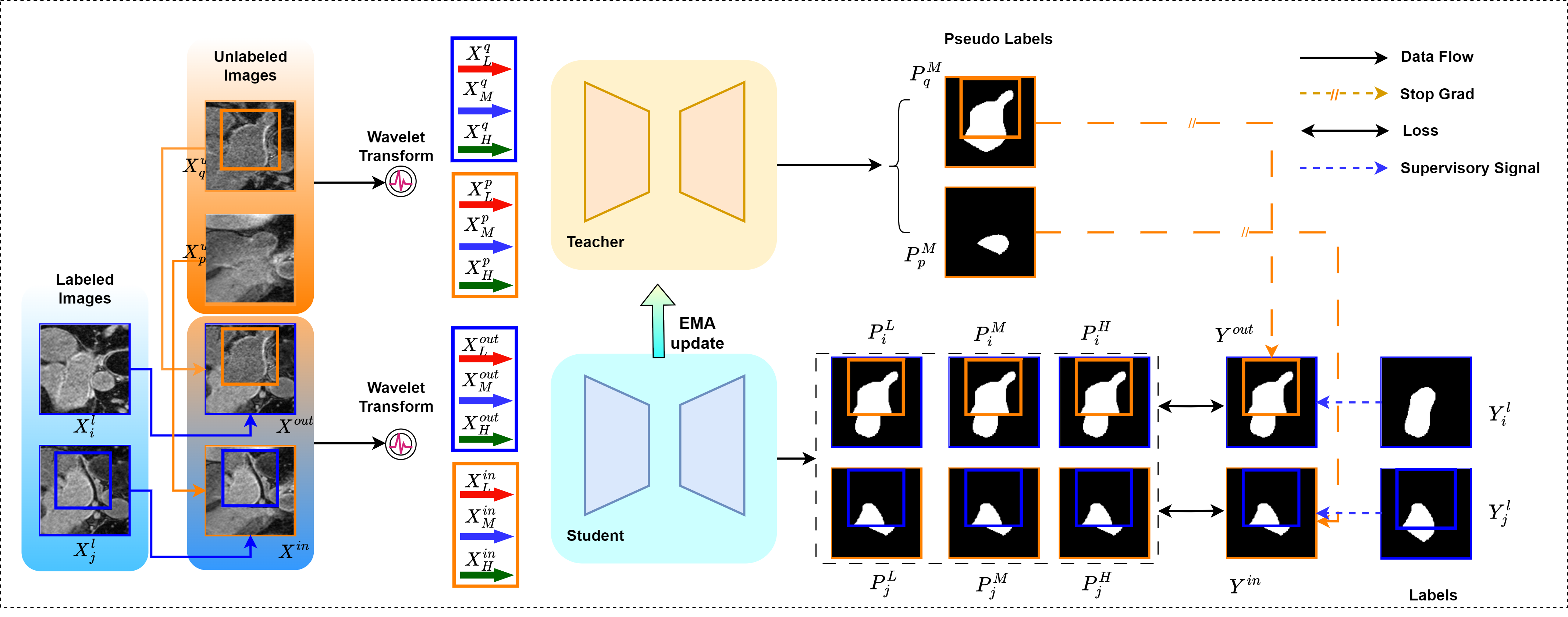}
    \caption{Overview of our WT-BCP framework for SSIM. The WT-BCP framework adopts a BCP strategy, merging two labeled and two unlabeled images to create mix images. The teacher uses unlabeled images to generate pseudo-labels. Ground truths and pseudo-labels are then mixed to produce mix labels. Student and Teacher model are XNet-Plus, as illustrated in Figure \ref{fig:net}}
    \label{fig:framework}
\end{figure*}

As shown in Figure \ref{fig:wt_abc}, 2D (3D) images are essentially 2D (3D) discrete nonstationary signals, containing different frequency ranges and spatial locations information. Wavelet transform can effectively preserve these information while decomposing them.
To address the above issues, we propose the Wavelet Transform based Bidirectional Copy-Paste (WT-BCP) framework. 
The specific structure is shown in Figure \ref{fig:framework}. We have made the following changes to the original BCP: (1) We added Wavelet Transform (WT) to deal with the output data of copy-paste to obtain the low frequency (LF) and high frequency (HF) information of input images. (2) We designed the Xnet-Plus model to receive image information fused with HF and LF and conduct more refined processing in the model, especially in terms of high-dimensional and low-dimensional feature fusion. (3) Using the Mean Teacher (MT) framework, we performed consistency supervision on the unlabeled data of the Xnet-Plus models of the Student network to optimize Xnet-Plus, and used the pseudo-labels of the Teacher and the ground truth labels to perform supervised learning on the Student network.
Our contributions are summarized as follows:
\begin{itemize}
    \item We propose a Bidirectional Copy-Paste framework based on Wavelet Transform for the first time, and the mixed HF and LF provide more image details and abstract semantics for network training.
    \item We propose the XNet-Plus model, which is used to process a multi-input and multi-output network that fuses HF, LF and the features of the original image. The multi-output fuses available HF and LF information for consistency learning, which can alleviate the learning bias caused by artificial perturbations.
    \item Our approach outperforms other SOTA SSMIS methods on three public medical image datasets, and it performs comparably or surpasses the fully supervised method.
\end{itemize}

\section{Related Works}
\label{sec:Related}

\subsection{Wavelet Transform in Deep Learning}
Wavelet transforms have become a go-to tool for manipulating time-series data and for enhancing images through various stages of processing. The Discrete Wavelet Transform (DWT) breaks down signals into a set of coefficients that represent different frequency bands and scales.
In \cite{fujieda2018wavelet}, wavelet transform was employed to exploit spectral information for texture classification and image annotation. A method that combines CNNs with wavelets for noise-resistant image classification was presented in \cite{li2021wavecnet}. Furthermore, \cite{yao2022wave} utilized wavelet transform to reduce the spatial resolution of key-value pairs in vision Transformers, thereby cutting down on computational expenses.
However, these methods still face limitations in their ability to decompose images into low-frequency and high-frequency components, and they struggle to capture the finer details of images effectively.
XNet~\cite{Zhou_2023_ICCV} and its enhanced version, XNetV2~\cite{zhou2024xnet}, pioneered the use of high-frequency (HF) and low-frequency (LF) components generated by wavelet transform for consistency loss calculation at the network's output. This innovative approach has yielded remarkable results in semi-supervised medical image segmentation.

\subsection{CutMix for Semi-Supervised Segmentation}
CutMix \cite{CutMix} enhances the accuracy of semi-supervised image segmentation by creating novel samples via the process of copying segments from one image and pasting them onto another. UCC \cite{fan2022ucc} addresses issues of distribution discrepancy and class imbalance by extracting portions from labeled images to use as the foreground and then overlaying them onto a different dataset. BCP \cite{bai2023bidirectional} introduces a technique for the two-way exchange of segments between labeled and unlabeled data, significantly narrowing the empirical distribution gap. This is achieved by prompting the unlabeled data to absorb a broad range of common features from both internal and external labeled data sources.
GSC-Seg \cite{jiang2024gradient} detects significant regions of labeled images by calculating network gradients and perform bidirectional copy-pasting with unlabeled images and proposes a gradient augmentation strategy to enhance the network’s gradient representation capability.
This approach fosters the unlabeled data's ability to grasp the contextual meaning of organs from the labeled data, which in turn boosts its capacity to recognize smaller organs. 


\section{Method}
\label{sec:Method}
We define the 3D volume of a medical image as $\textbf{X}\in\mathbb{R}^{W\times H\times L}$. The goal of  semi-supervised medical image segmentation is to predict the per-voxel label map $\widehat{\textbf{Y}}\in\{0,1, ..., K-1\}^{W\times H\times L}$, indicating where the background and the targets are in $\textbf{X}$. $K$ is the class number. Our training set $\mathcal{D}$ consists of $N$ labeled data and $M$ unlabeled data ($N\ll M$), expressed as two subsets: $\mathcal{D}=\mathcal{D}^l\cup \mathcal{D}^u$, where $\mathcal{D}^l = \{(\textbf{X}^l_i,\textbf{Y}^l_i)\}_{i=1}^N$ and $\mathcal{D}^u = \{\textbf{X}^u_i\}_{i=N+1}^{M+N}$.

\subsection{WT-BCP Framework}
This section details our proposed WT-BCP for SSIM as shown in Figure \ref{fig:framework}. Our framework consists of four parts: The Mean Teacher architecture, Wavelet Transform , XNet-Plus model and Bidirectional Copy-Paste.
The Mean Teacher architecture is composed of a student network and a teacher network, with the teacher network updated through a weighted combination of the student network parameters and an exponential moving average (EMA). The Semi-Supervised Learning (SSL) phase commences by initializing the students and the teacher with the pre-trained model, that is to say, we should first train the teacher network with the labeled data.

The overall pipeline of the proposed bidirectional copy-paste method is shown in Fig.~\ref{fig:framework}, in the Mean Teacher architecture. We randomly pick two unlabeled images $(\textbf{X}^u_p, \textbf{X}^u_q)$, and two labeled images $(\textbf{X}^l_i, \textbf{X}^l_j)$ from the training set. Then we copy-paste a random crop from $\textbf{X}^l_i$ (the background) onto $\textbf{X}^u_q$ (the foreground) to generate the mixed image $\textbf{X}^{out}$, and from $\textbf{X}^u_p$ (the background) onto $\textbf{X}^l_j$ (the foreground) to generate another mixed image $\textbf{X}^{in}$. Unlabeled images are able to learn comprehensive common semantics from labeled images from both \emph{inward} ($\textbf{X}^{in}$) and \emph{outward} ($\textbf{X}^{out}$) directions.

The mixture of labeled and unlabeled $(\textbf{X}^{in}, \textbf{X}^{out})$  and $(\textbf{X}^u_p, \textbf{X}^u_q)$ then pass through the WT module, and the features of LF (Low Frequency), HF (High Frequency) and the original images are outputs, which are represented by $\textbf{X}_{L}$, $\textbf{X}_{M}$ and $\textbf{X}_{H}$ respectively.
Both $\left (  \textbf{X}_{L}^{in}, \textbf{X}_{M}^{in},\textbf{X}_{H}^{in}\right ) $ and $\left (  \textbf{X}_{L}^{out}, \textbf{X}_{M}^{out},\textbf{X}_{H}^{out}\right ) $
at different frequencies are then fed into the Student network to predict segmentation masks, and these masks correspond to the outputs of the features at different frequencies in XNet-Plus. 
This will be introduced in detail in the next section.
Although both Teacher and Student networks use XNet-Plus model, in the Teacher network, we only use the output features of the original image $ \left ( P_{q}^{M}, P_{p}{M}\right )$.
The segmentation masks are supervised by bidirectional copy-pasting the predictions of the unlabeled images from the Teacher network and the ground truth labels.

\subsection{Bidirectional Copy-Paste Images}

We combine our framework with the SSMIS method BCP~\cite{ref2}. 
In the Semi-Supervised Learning (SSL) phase, we need to generate a zero-centered 
mask $\mathcal{M} \in {(0, 1)}^{W \times H \times D}$,
where 0 represents foreground and 1 represents background, the size of 
0 region is $\beta W \times \beta H \times \beta D$, referring to previous work \cite{ref2}, $\beta$ is set to $\frac{2}{3}$. Next, 
we use the mask to obtain mix images as the input of SSL as follows:
\begin{align}
&\textbf{X}^{in}=\textbf{X}^l_{j}\odot\mathcal{M}+\textbf{X}^u_{p}\odot\left(\textbf{1} - \mathcal{M} \right),\label{eq:mix1}\\
&\textbf{X}^{out}=\textbf{X}^u_{q}\odot\mathcal{M}+\textbf{X}^l_{i}\odot\left(\textbf{1} - \mathcal{M} \right),\label{eq:mix2}
\end{align}
where $\textbf{X}^l_{i}$, $\textbf{X}^l_{j}\in\mathcal{D}^l$, $i\neq j$, $\textbf{X}^u_{p}$, $\textbf{X}^u_{q}\in\mathcal{D}^u$, $p\neq q$, $\textbf{1}\in\{1\}^{W\times H\times L}$, and $\odot$ means element-wise multiplication. Two labeled and unlabeled images are adopted to keep the diversity of the input.

To obtain the pseudo labels, $x^u_p$ and $x^u_q$ are forwarded to the Teacher to get pseudo-labels $p^M_p$ and $p^M_q$. Due to passing through the XNet-Plus module, we avoid the LF (Low Frequency) and HF (High Frequency) parts with noise and directly select the branch of the raw image.
Then the mix labels are defined as follows:
\begin{align}
    \textbf{Y}^{in} &= \textbf{Y}^l_j \odot \mathcal{M} + \textbf{P}_q^M \odot (1-\mathcal{M}) \label{eq:Yin} \\
    \textbf{Y}^{out} &= \textbf{P}_p^M \odot \mathcal{M} + \textbf{Y}^l_i \odot (1-\mathcal{M}) \label{eq:Yout}
\end{align}
where $i \neq j$ and $p \neq q$, and $ \odot $ denotes element-wise multiplication.
$\textbf{Y}^{in}$ and $\textbf{Y}^{out}$ will be used as the supervision to supervise the Student network predictions of $\textbf{X}^{in}$ and $\textbf{X}^{out}$.


\begin{figure}
    \centering
    \includegraphics[width=\linewidth]{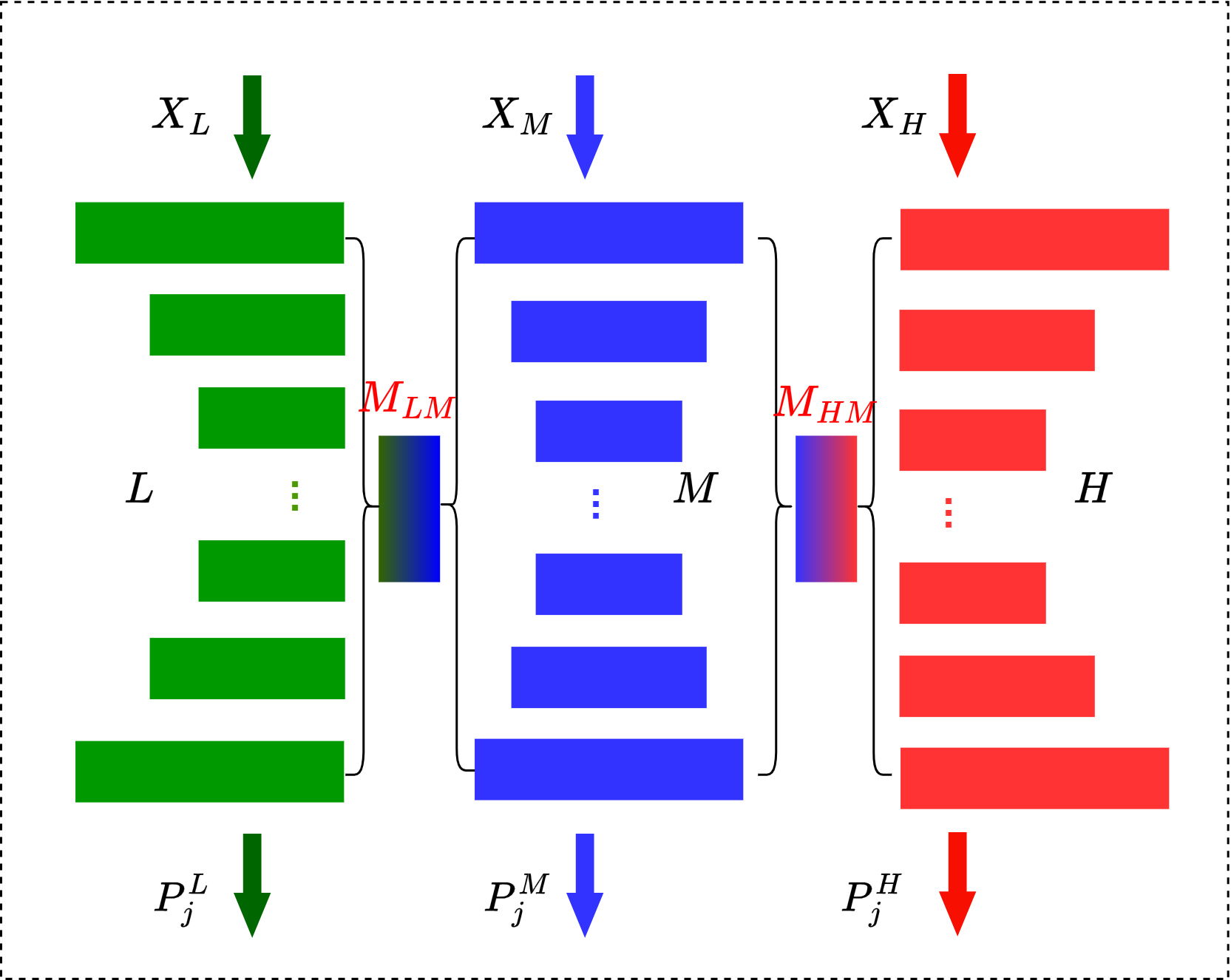}
    \caption{Overview of XNet-Plus model. XNet-Plus consists of main network $M$ , LF network $L$ and HF network $H$, and uses raw image $X_{M}$ , LF complementary fusion image $X_{L}$ and HF complementary fusion image $X_{H}$ as input.}
    \label{fig:net}
\end{figure}

\subsection{Wavelet Transform}
As shown in Figure \ref{fig:wt_abc}, we use wavelet transform to decompose raw images into LF, horizontal HF, vertical HF and diagonal HF components (LL, HL, LH and HH). They respectively save LF and different HF information of raw images. We represent LF images L with LF components and represent HF images H as the sum of HF components in different directions. L and H are defined as:
\begin{align}
    LF &= LL \\
    HF &= HL + LH + HH.
\end{align}
For semantic segmentation problem, accurate segmentation requires LF semantics (such as shape, color, etc.) and HF details (such as edges, textures, etc.).

\subsection{XNet-Plus}
\begin{figure}[!ht]
    \centering
    \includegraphics[width=\linewidth]{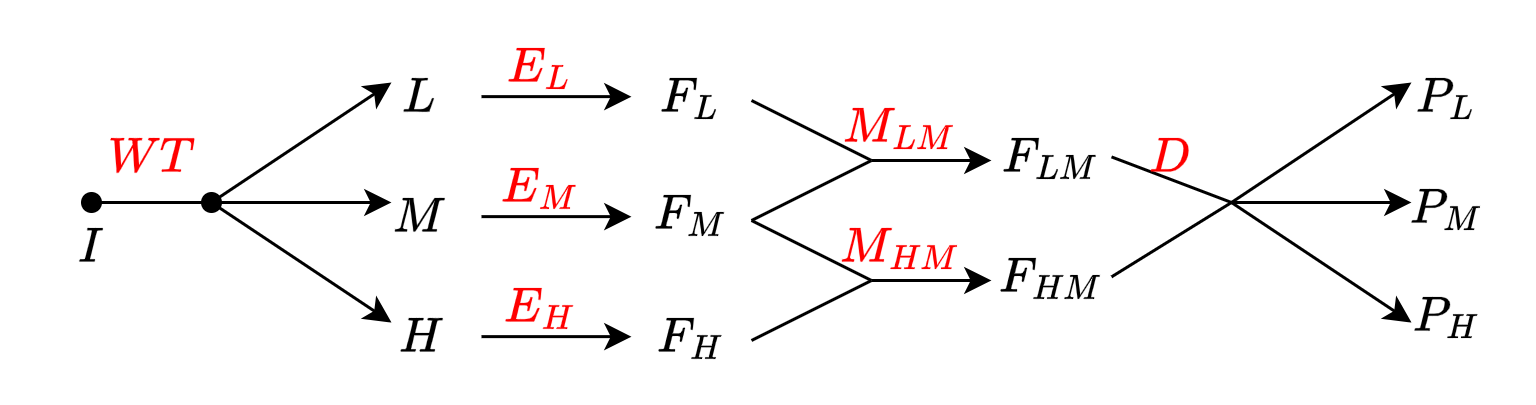}
    \caption{Topological flow chart of XNet-Plus data processing.}
    \label{fig:topo}
\end{figure}
As shown in Figure \ref{fig:net}, our XNet-Plus is a simple improvement on the UNet \cite{ronneberger2015u} and 3D UNet \cite{cciccek20163d}. The main modifications are increasing the number of inputs and outputs and adding the feature fusion module. The encoder-decoder parts are all based on traditional convolutions. 
The data flow topology diagram is illustrated in Figure \ref{fig:topo}, the raw image $I$ consists of Low Frequency (LF) features $F_{L}$, High Frequency (HF) features $F_{H}$, LF additive noise $N_{L}$, and HF additive noise $N_{H}$. Therefore, $I$ can be defined as:
\begin{equation}
    I = M = F_{L} + F_{H} + N_{L} + N_{H}
\end{equation}
Although LF and HF information can be fused into complete information in fusion module, the raw image may still contain useful but unappreciated information, therefore, we also take the original image as the input of the network and denote it with the symbol $M$.

Wavelet transform WT can decouple image I to generate LF image L and HF image H:
\begin{equation}
        L,H = WT\left ( I \right ), L = F_{L} + N_{L}, H = F_{H} + N_{H}.
\end{equation}
Raw image $M$, LF and HF encoders $E_{M}$ $E_{L}$, $E_{H}$ extract $F_{M}$, $F_{L}$ and $F_{H}$ from $M$, L and H, respectively:
\begin{equation}
    F_{M} = E_{M}\left ( M \right ), \\
    F_{L} = E_{L}\left ( L \right ), \\ 
    F_{H} = E_{H}\left ( H \right )
\end{equation}
We perform fusion operations on the features by concatenation.
Fusion modules $M_{LM}$ and $M_{HM}$ fuse $ \left( F_{L} , F_{M} \right) $ and $ \left( F_{H} , F_{M} \right) $ to acquire the complete features $F_{LM}$ and $F_{HM}$:
\begin{align}
    &F_{LM} = M_{LM} \left ( F_{L}, F_{M} \right ) \\
    &F_{HM} = M_{HM} \left ( F_{H}, F_{M} \right )
\end{align}
The segmentation predictions of $M$, LF and HF branches are defined as:
\begin{equation}
    P_{M}, P_{L}, P_{H} = D \left( F_{LM}, F_{HM} \right )
\end{equation}
where $P_{M}$, $P_{L}$ and $P_{H}$ represent $M$, LF and HF segmentation predictions, $D$ represents tri-branch decoder.


\subsection{Loss Function}
The labeled and unlabeled images first undergo BCP and then WT operations; their outputs $\left( X_L^{out}, X_M^{out}, X_H^{out} \right)$ and $\left( X_L^{in}, X_M^{in}, X_H^{in} \right)$ will serve as the input for the Student network.
We use $\alpha$ to control the contribution of unlabeled image pixels to the loss function. The loss functions for $\textbf{P}^{k}_{j}$ and $\textbf{P}^{k}_{i}$ , $k \in \left \{ L, M, H \right \}  $  are computed respectively by
\begin{align}
\label{eq:loss}
\mathcal{L}^{in}=\mathcal{L}_\textit{seg}&\left(\textbf{P}^{k}_{j},\textbf{Y}^{in}\right)\odot\mathcal{M}+\\\nonumber
&\alpha\mathcal{L}_\textit{seg}\left(\textbf{P}^{k}_{j},\textbf{Y}^{in}\right)\odot\left(\textbf{1}-\mathcal{M}\right),\\
\mathcal{L}^{out}=\mathcal{L}_\textit{seg}&\left(\textbf{P}^{k}_{i},\textbf{Y}^{out}\right)\odot\left(\textbf{1}-\mathcal{M}\right)+\\\nonumber
&\alpha\mathcal{L}_\textit{seg}\left(\textbf{P}^{k}_{i},\textbf{Y}^{out}\right)\odot\mathcal{M},
\end{align}
where $\mathcal{L}_{seg}$ is the linear combination of Dice loss and Cross-entropy loss, $\alpha$ is set to 0.5 referring to previous work ~\cite{ref2}.
$\textbf{Y}^{in}$ and $\textbf{Y}^{out}$ are computed in Formula \ref{eq:Yin} and Formula \ref{eq:Yout}.
$\textbf{P}^{k}_{i}$ and $\textbf{P}^{k}_{j}$ are computed by:
\begin{align}
    &\textbf{P}^{k}_{j}=\mathcal{F}_s(WT\left(\textbf{X}^{in}\right);\mathbf{\Theta}_s) \\
    &\textbf{P}^{k}_{i}=\mathcal{F}_s(WT\left( \textbf{X}^{out} \right );\mathbf{\Theta}_s)
\end{align}
At each iteration we update the parameters $\mathbf{\Theta}_{s}$ in Student network by stochastic gradient descent with the loss function:
\begin{equation}
\mathcal{L}_\textit{all}=\mathcal{L}^{in}+\mathcal{L}^{out} +\beta\mathcal{L}_{con}
\end{equation}
Referencing the previous work \cite{chen2021semi}, we calculate the consistency loss using the Dice loss $\mathcal{L}_{dice}$. $\beta$ is a hyperparameter, and by default, it is chosen to be 1.
We calculate the triple output complementary consistency loss of XNet-Plus in the Student network.
\begin{align}
    &\mathcal{L}_{con} = \mathcal{L}^{M,L}_{con} + \mathcal{L}^{M,H}_{con} \\
    &\mathcal{L}^{M,L}_{con} =\mathcal{L}_{dice} \left( \textbf{P}^{M} , \textbf{P}^{L} \right) \\
    &\mathcal{L}^{M,H}_{con} = \mathcal{L}_{dice} \left( \textbf{P}^{M} , \textbf{P}^{H} \right)
\end{align}
For the two input groups in the Student network, the consistency loss needs to be calculated for both, we use $\mathcal{L}_{con}$ and $\textbf{P}^{k}$ , $k \in \left \{ L, M, H \right \}$ for simplified notation.

Teacher network parameters $\mathbf{\Theta}_{t}^{(k+1)}$ at the $\left( k+1\right)$th iteration are updated:
\begin{equation}
    \mathbf{\Theta}_{t}^{(k+1)} = \lambda \mathbf{\Theta}_{t}^{(k)} + \left( 1 - \lambda \right) \mathbf{\Theta}_{s}^{(k)}
\end{equation}
where $\lambda$ is the smoothing coefficient parameter.

\subsection{Testing Phase}
In the testing stage, given a testing image $\textbf{X}_{test}$, we obtain the probability map by:
    $\textbf{Q}_{test}=\mathcal{F}(\mathcal{WT}\left( \textbf{X}_{test} \right ); \widehat{\mathbf{\Theta}}_s)$,
where $\widehat{\mathbf{\Theta}}_s$ are the well-trained Student network parameters. The final label map can be easily determined by $\textbf{Q}_{test}$.

\section{Experiments}
\label{sec:Exper}
\subsection{Dataset}
\noindent\textbf{ACDC dataset}. ACDC \cite{ACDCdataset} dataset is a four-class (\emph{i.e.} background, right ventricle, left ventricle and myocardium) segmentation dataset, containing 100 patients' scans. The data split \cite{ssl4mis} is fixed with 70, 10, and 20 patients' scans for training, validation, and testing.

\noindent\textbf{LA dataset}. Atrial Segmentation Challenge \cite{LAdataset} dataset includes 100 3D gadolinium-enhanced magnetic resonance image scans (GE-MRIs) with labels. We strictly follow the setting used in SSNet \cite{SSNet}, DTC \cite{Luo2021DTC} and UA-MT \cite{UAMT}.

\noindent\textbf{Pancreas-NIH}. Pancreas-NIH \cite{NIHPancreas} dataset contains 82 contrast-enhanced abdominal CT volumes which are manually delineated. For fair comparison, we follow the setting in CoraNet \cite{CoraNet}.

\begin{table}[!h]
\caption{Comparisons with state-of-the-art semi-supervised segmentation methods on the ACDC dataset.}
\label{tab:res-acdc}
\resizebox{\columnwidth}{!}{%
\begin{tabular}{l|cc|llll}
\hline
\multicolumn{1}{c|}{\multirow{2}{*}{Method}} & \multicolumn{2}{c|}{Scans used} & \multicolumn{4}{c}{Metrics} \\ \cline{2-7} 
\multicolumn{1}{c|}{} & \multicolumn{1}{l}{Labeled} & \multicolumn{1}{l|}{Unlabeled} & Dice↑ & Jaccard↑ & 95HD↓ & ASD↓ \\ \hline
U-Net & \multicolumn{1}{l}{3(5\%)} & \multicolumn{1}{l|}{0} & 47.83 & 37.01 & 31.16 & 12.62 \\
U-Net & \multicolumn{1}{l}{7(10\%)} & \multicolumn{1}{l|}{0} & 79.41 & 68.11 & 9.35 & 2.7 \\
U-Net & \multicolumn{1}{l}{70(All)} & \multicolumn{1}{l|}{0} & 91.44 & 84.59 & 4.3 & 0.99 \\ \hline
UA-MT & \multirow{8}{*}{3(5\%)} & \multirow{8}{*}{67(95\%)} & 46.04 & 35.97 & 20.08 & 7.75 \\
SASSNet &  &  & 57.77 & 46.14 & 20.05 & 6.06 \\
DCT &  &  & 56.9 & 45.67 & 23.36 & 7.39 \\
URPC &  &  & 55.87 & 44.64 & 13.6 & 3.74 \\
MC-Net &  &  & 62.85 & 52.29 & 7.62 & 2.33 \\
SS-Net &  &  & 65.83 & 55.38 & 6.67 & 2.28 \\
BCP &  &  & 87.59 & 78.67 & 1.9 & 0.67 \\
XnetV2 &  &  & 87.94 & 78.83 & 1.82 & 0.53 \\
Ours &  &  & \textbf{88.24} & \textbf{79.03} & \textbf{1.7} & \textbf{0.58} \\ \hline
UA-MT & \multirow{8}{*}{7(10\%)} & \multirow{8}{*}{63(90\%)} & 81.65 & 70.64 & 6.88 & 2.02 \\
SASSNet &  &  & 84.5 & 74.34 & 5.42 & 1.86 \\
DCT &  &  & 84.29 & 73.92 & 12.81 & 4.01 \\
URPC &  &  & 83.1 & 72.41 & 4.84 & 1.53 \\
MC-Net &  &  & 86.44 & 77.04 & 5.5 & 1.84 \\
SS-Net &  &  & 86.78 & 77.67 & 6.07 & 1.4 \\
BCP &  &  & 88.84 & 80.62 & 3.98 & 1.17 \\
XNetV2 &  &  & 89.35 & 81.59 & 2.66 & 0.99 \\
Ours &  &  & \textbf{89.97} & \textbf{82.32} & \textbf{1.51} & \textbf{0.52} \\ \hline
\end{tabular}%
}
\end{table}

\subsection{Implementation details}
In this work, all our experiments were conducted in parallel on four NVIDIA V100 GPUs (each with 32GB memory) using the PyTorch framework.
We performed random cropping on these data. For LA, it was cropped to $112\times112\times80$. For Pancrease-NIH, it was $90\times90\times90$. And for ACDC, it was $256\times256$. 

\subsection{Evaluation Metrics} We choose four evaluation metrics: \textit{Dice Score} (\%), \textit{Jaccard Score} (\%), \textit{95\% Hausdorff Distance (95HD) in voxel} and \textit{Average Surface Distance (ASD) in voxel}. Given two object regions, Dice and Jaccard mainly compute the percentage of overlap between them, ASD computes the average distance between their boundaries, and 95HD measures the closest point distance between them.

\subsection{Quantitative Comparison}
As shown in Table \ref{tab:res-acdc} and Table \ref{tab:res-la},  we compare our method with various competitor methods on the ACDC and LA datasets: UA-MT \cite{UAMT}, SASSNet \cite{SASSNet}, DTC \cite{Luo2021DTC}, URPC \cite{URPC}, MC-Net \cite{MCNet}, SS-Net \cite{SSNet}, XNetV2 \cite{zhou2024xnet} and BCP \cite{ref2}, and conduct semisupervised experiments at different labeling ratios.

Our algorithm surpasses the aforementioned models in performance metrics. Consequently, during training, WT-BCP not only efficiently transfers labels from labeled data to unlabeled data but also enhances accuracy in boundary assessment. The model is capable of detecting more nuanced semantic changes within boundary areas or voxels.

We conducted experiments on the Pancreas-NIH dataset using a 20$\%$ labeling ratio. The pancreas, situated deep within the abdominal cavity, exhibits significant variation in size, position, and form. In Table \ref{tab:res-pct}, we compared our approach with V-Net \cite{VNet}, DAN \cite{DAN}, ADVNET \cite{ADVNet}, UA-MT \cite{UAMT}, SASSNet \cite{SASSNet}, DTC \cite{Luo2021DTC}, CoraNet \cite{CoraNet}, XNetV2 \cite{zhou2024xnet} and BCP \cite{ref2}. The quantitative outcomes demonstrated that WT-BCP outperformed all other cutting-edge techniques, indicating that the WT-BCP framework possesses strong generalization and stability.

\begin{table}[]
\caption{Comparisons with state-of-the-art semi-supervised segmentation methods on the LA dataset.}
\label{tab:res-la}
\resizebox{\columnwidth}{!}{%
\begin{tabular}{l|ll|llll}
\hline
\multirow{2}{*}{Method} & \multicolumn{2}{l|}{Scans used} & \multicolumn{4}{l}{Metrics} \\ \cline{2-7} 
 & Labeled & Unlabeled & Dice↑ & Jaccard↑ & 95HD↓ & ASD↓ \\ \hline
V-Net & 4(5\%) & 0 & 52.55 & 39.6 & 47.05 & 9.87 \\
V-Net & 8(10\%) & 0 & 82.74 & 71.72 & 13.35 & 3.26 \\
V-Net & 80(All) & 0 & 91.47 & 84.36 & 5.48 & 1.51 \\ \hline
UA-MT & \multirow{8}{*}{4(5\%)} & \multirow{8}{*}{76(95\%)} & 82.26 & 70.98 & 13.71 & 3.82 \\
SASSNet &  &  & 81.6 & 69.63 & 16.16 & 3.58 \\
DCT &  &  & 81.25 & 69.33 & 14.9 & 3.39 \\
URPC &  &  & 82.48 & 71.35 & 14.65 & 3.65 \\
MC-Net &  &  & 83.59 & 72.36 & 14.07 & 2.7 \\
SS-Net &  &  & 86.33 & 76.15 & 9.97 & 2.31 \\
BCP &  &  & 88.02 & 78.72 & 7.9 & 2.15 \\
XNetV2 &  &  & 88.55 & 78.91 & 7.82 & 2.07 \\
Ours &  &  & \textbf{89.24} & \textbf{79.07} & \textbf{7.03} & \textbf{1.89} \\ \hline
UA-MT & \multirow{8}{*}{8(10\%)} & \multirow{8}{*}{72(90\%)} & 87.79 & 78.39 & 8.68 & 2.12 \\
SASSNet &  &  & 87.54 & 78.05 & 9.84 & 2.59 \\
DCT &  &  & 87.51 & 78.17 & 8.23 & 2.36 \\
URPC &  &  & 86.92 & 77.03 & 11.13 & 2.28 \\
MC-Net &  &  & 87.62 & 78.25 & 10.03 & 1.82 \\
SS-Net &  &  & 88.55 & 79.62 & 7.49 & 1.9 \\
BCP &  &  & 89.62 & 81.31 & 6.81 & 1.76 \\
XNetV2 &  &  & 90.73 & 82.81 & 5.68 & 1.57 \\
Ours &  &  & \textbf{91.01} & \textbf{83.84} & \textbf{5.02} & \textbf{1.55} \\ \hline
\end{tabular}%
}
\end{table}

\begin{table}[]
\caption{Comparisons with state-of-the-art semi-supervised segmentation methods on the Pancrease-NIH dataset.}
\label{tab:res-pct}
\resizebox{\columnwidth}{!}{%
\begin{tabular}{l|ll|llll}
\hline
\multirow{2}{*}{Method} & \multicolumn{2}{l|}{Scans used} & \multicolumn{4}{l}{Metrics} \\ \cline{2-7} 
 & Labeled & Unlabeled & Dice↑ & Jaccard↑ & 95HD↓ & ASD↓ \\ \hline
V-Net & \multirow{9}{*}{12(20\%)} & \multirow{9}{*}{50(80\%)} & 69.96 & 55.55 & 14.27 & 1.64 \\
DAN &  &  & 76.74 & 63.29 & 11.13 & 2.97 \\
ADVNet &  &  & 75.31 & 61.73 & 11.72 & 3.88 \\
UA-MT &  &  & 77.26 & 63.82 & 11.9 & 3.06 \\
SASSNet &  &  & 77.66 & 64.08 & 10.93 & 3.05 \\
DTC &  &  & 78.27 & 64.75 & 8.36 & 2.25 \\
CoraNet &  &  & 79.67 & 66.69 & 7.59 & 1.89 \\
BCP &  &  & 82.91 & 70.97 & 6.43 & 2.25 \\
XNetV2 &  &  & 83.25 & 71.44 & 5.7 & 1.83 \\
\textbf{Ours} &  &  & \textbf{84.09} & \textbf{72.87} & \textbf{4.99} & \textbf{1.3} \\ \hline
\end{tabular}%
}
\end{table}

\subsection{Ablation Study}
\begin{table}[]
\caption{Ablation Study of different frequencies on ACDC}
\label{tab:res-ab}
\resizebox{\columnwidth}{!}{%
\begin{tabular}{l|cc|llll}
\hline
\multicolumn{1}{c|}{\multirow{2}{*}{Method}} & \multicolumn{2}{c|}{Scans used} & \multicolumn{4}{c}{Metrics} \\ \cline{2-7} 
\multicolumn{1}{c|}{} & \multicolumn{1}{l}{Labeled} & \multicolumn{1}{l|}{Unlabeled} & Dice↑ & Jaccard↑ & 95HD↓ & ASD↓ \\ \hline
M & \multirow{4}{*}{7(10\%)} & \multirow{4}{*}{63(90\%)} & 87.21 & 77.38 & 2.64 & 0.82 \\
M + LF &  &  & 88.47 & 78.29 & 2.07 & 0.69 \\
M + HF &  &  & 89.04 & 79.53 & 1.94 & 0.65 \\
M + LF + HF &  &  & \textbf{89.97} & \textbf{82.32} & \textbf{1.51} & \textbf{0.52} \\ \hline
\end{tabular}%
}
\end{table}

We conducted ablation experiments on the ACDC dataset and used 10$\%$ of the labeled data to perform combination and comparison of different frequencies of data for the Xnet-plus model in our proposed WT-BCP. Four groups of comparative experiments were designed to discuss whether it is necessary to fuse data of different frequencies. Among them, the raw image is denoted as M, the low frequency as LF, and the high frequency as HF.
In the XNet-plus model, Table \ref{tab:res-ab} reveals that as more frequency information is fused, the segmentation performance gradually improves. 
$+$ represents information fusion.
Because LF itself contains semantic information such as shape, color, etc., while HF contains more detailed information such as edges, textures, etc.

\section{Conclusion}
\label{sec:Conclusion}
We propose a Wavelet Transform based Bidirectional Copy-Paste SSMIS method, named WT-BCP. 
We apply the Wavelet Transform (WT) after bidirectionally copying and pasting to better ensure the consistency and distribution balance of semi-supervised training data. 
The output of WT contains three parts, namely the raw image, LF, and HF. 
Moreover, we have designed the XNet-plus model that is common to both the Teacher and Student networks. 
The three output parts of WT can perform feature fusion better, enhance the extraction capabilities of semantics and details, and improve the segmentation performance.
Our experiments on the ACDC, LA, and Pancreas-NIH datasets demonstrate the superiority of the proposed WT-BCP method, and we achieve significant performance improvements on all four segmentation metrics.


\bibliographystyle{abbrv}
\bibliography{icme2025references}

\begin{thebibliography}{10}

\bibitem{arazo2020pseudo}
E.~Arazo, D.~Ortego, et~al.
\newblock Pseudo-labeling and confirmation bias in deep semi-supervised
  learning.
\newblock In {\em 2020 IJCNN Proceedings}.

\bibitem{bai2023bidirectional}
Y.~Bai, D.~Chen, et~al.
\newblock Bidirectional copy-paste for semi-supervised medical image
  segmentation.
\newblock In {\em Proceedings of CVPR}.

\bibitem{ref2}
Y.~Bai, D.~Chen, Q.~Li, W.~Shen, and Y.~Wang.
\newblock Bidirectional copy-paste for semi-supervised medical image
  segmentation.
\newblock In {\em Proceedings of CVPR}.

\bibitem{ACDCdataset}
O.~Bernard and A.~Lalande.
\newblock Deep learning techniques for automatic {MRI} cardiac multi-structures
  segmentation and diagnosis: Is the problem solved?
\newblock {\em {IEEE} Trans. Medical Imaging}.

\bibitem{chen2021semi}
X.~Chen, Y.~Yuan, G.~Zeng, and J.~Wang.
\newblock Semi-supervised semantic segmentation with cross pseudo supervision.
\newblock In {\em Proceedings of CVPR}.

\bibitem{cciccek20163d}
{\"O}.~{\c{C}}i{\c{c}}ek, A.~Abdulkadir, et~al.
\newblock 3d u-net: learning dense volumetric segmentation from sparse
  annotation.
\newblock In {\em MICCAI 2016}.

\bibitem{Dou2020unpaired}
Q.~Dou, Q.~Liu, P.~Heng, and B.~Glocker.
\newblock Unpaired multi-modal segmentation via knowledge distillation.
\newblock {\em {IEEE} Trans. Medical Imaging}.

\bibitem{fan2022ucc}
J.~Fan, B.~Gao, H.~Jin, and L.~Jiang.
\newblock Ucc: Uncertainty guided cross-head co-training for semi-supervised
  semantic segmentation.
\newblock In {\em Proceedings of CVPR}.

\bibitem{fujieda2018wavelet}
S.~Fujieda, K.~Takayama, and T.~Hachisuka.
\newblock Wavelet convolutional neural networks.
\newblock {\em arXiv preprint arXiv:1805.08620}.

\bibitem{jiang2024gradient}
Y.~Jiang, G.~Zhu, Y.~Ding, Z.~Qin, and M.~Pang.
\newblock Gradient saliency-aware cutmix for semi-supervised medical image
  segmentation.
\newblock In {\em 2024 IEEE International Conference on Multimedia and Expo
  (ICME)}.

\bibitem{li2021wavecnet}
Q.~Li, L.~Shen, S.~Guo, and Z.~Lai.
\newblock {WaveCNet}: Wavelet integrated {CNNs} to suppress aliasing effect for
  noise-robust image classification.
\newblock {\em IEEE Transactions on Image Processing}.

\bibitem{SASSNet}
S.~Li, C.~Zhang, and X.~He.
\newblock Shape-aware semi-supervised 3d semantic segmentation for medical
  images.
\newblock In {\em Proc. MICCAI}, 2020.

\bibitem{Li2018hdenseunet}
X.~Li, H.~Chen, X.~Qi, Q.~Dou, C.~Fu, and P.~Heng.
\newblock H-denseunet: Hybrid densely connected unet for liver and tumor
  segmentation from {CT} volumes.
\newblock {\em {IEEE} Trans. Medical Imaging}.

\bibitem{ssl4mis}
X.~Luo.
\newblock {SSL4MIS}.
\newblock \url{https://github.com/HiLab-git/SSL4MIS}, 2020.

\bibitem{Luo2021DTC}
X.~Luo, J.~Chen, T.~Song, and G.~Wang.
\newblock Semi-supervised medical image segmentation through dual-task
  consistency.
\newblock In {\em Proc. AAAI}.

\bibitem{URPC}
X.~Luo, W.~Liao, et~al.
\newblock Efficient semi-supervised gross target volume of nasopharyngeal
  carcinoma segmentation via uncertainty rectified pyramid consistency.
\newblock In {\em {MICCAI} 2021}.

\bibitem{VNet}
F.~Milletari, N.~Navab, and S.~Ahmadi.
\newblock V-net: Fully convolutional neural networks for volumetric medical
  image segmentation.
\newblock In {\em Fourth International Conference on 3D Vision}.

\bibitem{ronneberger2015u}
O.~Ronneberger, P.~Fischer, and T.~Brox.
\newblock U-net: Convolutional networks for biomedical image segmentation.
\newblock In {\em MICCAI 2015}.

\bibitem{NIHPancreas}
H.~R. Roth, L.~Lu, et~al.
\newblock Deeporgan: Multi-level deep convolutional networks for automated
  pancreas segmentation.
\newblock In {\em MICCAI 2015 - 18th International Conference}.

\bibitem{CoraNet}
Y.~Shi, J.~Zhang, T.~Ling, J.~Lu, Y.~Zheng, Q.~Yu, L.~Qi, and Y.~Gao.
\newblock Inconsistency-aware uncertainty estimation for semi-supervised
  medical image segmentation.
\newblock {\em {IEEE} Trans. Medical Imaging}.

\bibitem{ref1}
A.~Tarvainen and H.~Valpola.
\newblock Mean teachers are better role models: Weight-averaged consistency
  targets improve semi-supervised deep learning results.
\newblock {\em Advances in neural information processing systems}.

\bibitem{ADVNet}
T.~Vu, H.~Jain, M.~Bucher, M.~Cord, and P.~P{\'{e}}rez.
\newblock {ADVENT}: adversarial entropy minimization for domain adaptation in
  semantic segmentation.
\newblock In {\em {IEEE} Conference on Computer Vision and Pattern
  Recognition}.

\bibitem{Wang2022separated}
J.~Wang, X.~Li, Y.~Han, J.~Qin, L.~Wang, and Q.~Zhou.
\newblock Separated contrastive learning for organ-at-risk and
  gross-tumor-volume segmentation with limited annotation.
\newblock In {\em Proc. AAAI}.

\bibitem{ref4}
Y.~Wang, B.~Xiao, X.~Bi, W.~Li, and X.~Gao.
\newblock Mcf: Mutual correction framework for semi-supervised medical image
  segmentation.
\newblock In {\em Proceedings of CVPR}.

\bibitem{Wang2019abdominal}
Y.~Wang, Y.~Zhou, W.~Shen, S.~Park, E.~K. Fishman, and A.~L. Yuille.
\newblock Abdominal multi-organ segmentation with organ-attention networks and
  statistical fusion.
\newblock {\em Medical image analysis}.

\bibitem{SSNet}
Y.~Wu, Z.~Wu, Q.~Wu, Z.~Ge, and J.~Cai.
\newblock Exploring smoothness and class-separation for semi-supervised medical
  image segmentation.
\newblock {\em CoRR}.

\bibitem{MCNet}
Y.~Wu, M.~Xu, et~al.
\newblock Semi-supervised left atrium segmentation with mutual consistency
  training.
\newblock In {\em MICCAI 2021}.

\bibitem{LAdataset}
Z.~Xiong and Q.~Xia.
\newblock A global benchmark of algorithms for segmenting the left atrium from
  late gadolinium-enhanced cardiac magnetic resonance imaging.
\newblock {\em Medical Image Anal.}

\bibitem{yao2022wave}
T.~Yao, Y.~Pan, Y.~Li, C.-W. Ngo, and T.~Mei.
\newblock {Wave-ViT}: Unifying wavelet and {Transformers} for visual
  representation learning.
\newblock In {\em European Conference on Computer Vision}.

\bibitem{UAMT}
L.~Yu, S.~Wang, et~al.
\newblock Uncertainty-aware self-ensembling model for semi-supervised 3d left
  atrium segmentation.
\newblock In {\em Proc. MICCAI}.

\bibitem{CutMix}
S.~Yun, D.~Han, S.~Chun, S.~J. Oh, Y.~Yoo, and J.~Choe.
\newblock Cutmix: Regularization strategy to train strong classifiers with
  localizable features.
\newblock In {\em Proc. ICCV}.

\bibitem{DAN}
Y.~Zhang, L.~Yang, et~al.
\newblock Deep adversarial networks for biomedical image segmentation utilizing
  unannotated images.
\newblock In {\em MICCAI 2017}.

\bibitem{Zhao20213d}
T.~Zhao, K.~Cao, et~al.
\newblock 3d graph anatomy geometry-integrated network for pancreatic mass
  segmentation, diagnosis, and quantitative patient management.
\newblock In {\em Proc. CVPR}.

\bibitem{Zhou_2023_ICCV}
Y.~Zhou, J.~Huang, et~al.
\newblock Xnet: Wavelet-based low and high frequency fusion networks for fully-
  and semi-supervised semantic segmentation of biomedical images.
\newblock In {\em Proceedings of the IEEE/CVF International Conference on
  Computer Vision (ICCV)}.

\bibitem{zhou2024xnet}
Y.~Zhou, L.~Li, et~al.
\newblock Xnet v2: Fewer limitations, better results and greater universality.
\newblock {\em arXiv preprint arXiv:2409.00947}.

\end{thebibliography}

\end{document}